# Turning Speech Into Scripts

## Manny Rayner, Beth Ann Hockey, Frankie James

RIACS
Mail Stop 19–39
NASA Ames Research Center
Moffett Field, CA 94035–1000
{mrayner, bahockey, fjames}@riacs.edu

### Abstract

We describe an architecture for implementing spoken natural language dialogue interfaces to semi-autonomous systems, in which the central idea is to transform the input speech signal through successive levels of representation corresponding roughly to linguistic knowledge, dialogue knowledge, and domain knowledge. The final representation is an executable program in a simple scripting language equivalent to a subset of CSHELL. At each stage of the translation process, an input is transformed into an output, producing as a by-product a "meta-output" which describes the nature of the transformation performed. We show how consistent use of the output/meta-output distinction permits a simple and perspicuous treatment of apparently diverse topics including resolution of pronouns, correction of user misconceptions, and optimization of scripts. The methods described have been concretely realized in a prototype speech interface to a simulation of the Personal Satellite Assistant.

## Introduction

The basic task we consider in this paper is that of using spoken language to give commands to a semi-autonomous robot or other similar system. As evidence of the importance of this task in the NLP community note that the early, influential system SHRDLU [17] was intended to address just this type of problem. More recent work on spoken language interfaces to semi-autonomous robots include SRI's Flakey robot [4] and NCARAI's InterBOT project [9] [10]. A number of other systems have addressed part of the task. CommandTalk [6], Circuit Fix-It Shop [12] and TRAINS-96 [14] [15] are spoken language systems but they interface to simulation or help facilities rather than semi-autonomous agents. Jack's MOOse Lodge [1] takes text rather than speech as natural language input and the avatars being controlled are not semi-autonomous. Other researchers have considered particular aspects of the problem such as accounting for various aspects of actions [16] [11]. In most of this and other related work the treatment is some variant of the following. If there is a speech interface, the input speech signal is converted into text. Text either from the recognizer or directly input by the user is then converted into some kind of logical formula, which abstractly represents the user's intended command; this formula is then fed into a command interpreter, which executes the command.

We do not think the standard picture is in essence incorrect, but we do believe that, as it stands, it is in need of some modification. This paper will in particular make three points. First, we suggest that the output representation should not be regarded as a logical expression, but rather as a program in some kind of scripting language. Second, we argue that it is not merely the case that the process of converting the input signal to the final representation can sometimes go wrong; rather, this is the normal course of events, and the interpretation process should be organized with that assumption in mind. Third, we claim, perhaps surprisingly, that the first and second points are related.

We describe an architecture which addresses the issues outlined above, and which has been used to implement a prototype speech interface to a simulated semi-autonomous robot intended for deployment on the International Space Station, and present illustrative examples of interactions with the system.

## Theoretical Ideas

### Scripts vs. Logical Forms

Let's first look in a little more detail at the question of what the output representation should be. In practice, there seem to be two main choices: atheoretical representations, or some kind of logic.

Now, logic is indeed an excellent way to think about representing static relationships like database queries, but it is much less clear that it is a good way to represent *commands*. In real life, when people wish to give a command to a computer, they usually do so via its operating system; a complex command is an expression in a scripting language like CSHELL, Perl, or Visual Basic. These languages are related to logical formalisms, but cannot be mapped onto them in a simple way. Here are some of the obvious differences:

- A scripting language is essentially imperative, rather than relational.
- The notion of temporal sequence is fundamental to the language. "Do P and then Q" is not the same as "Make the goals P and Q true"; it is explicitly stated that P is to be done first. Similarly, "For each X in the list (A B C), do P(X)" is not the same as "For all X, make P(X) true"; once again, the scripting language defines an order, but not the logical language.
- Scripting languages assume that commands do not always succeed. For example, UNIX-based scripting

languages like CSHELL provide each script with the three predefined streams `stdin`, `stdout` and `stderr`. Input is read from `stdin` and written to `stdout`; error messages, warnings and other comments are sent to `stderr`.

We do not think that these properties of scripting language are accidental. They have evolved as the result of strong selectional pressure from real users with real-world tasks that need to be carried out, and represent a competitive way to meet said user's needs. We consequently think it is worth taking seriously the idea that a target representation produced by a spoken language interface should share many of these properties.

## Fallible Interpretation: Outputs and Meta-outputs

We now move on to the question of modelling the interpretation process, that is to say the process which converts the input (speech) signal to the output (executable) representation. As already indicated, we think it is important to realize that interpretation is a process which, like any other process, may succeed more or less well in achieving its intended goals. Users may express themselves unclearly or incompletely, or the system may more or less seriously fail to understand exactly what they mean. A good interpretation architecture will keep these considerations in mind.

Taking our lead from the description of scripting languages sketched above, we adapt the notion of the "error stream" to the interpretation process. In the course of interpreting an utterance, the system translates it into successively "deeper" levels of representation. Each translation step has not only an input (the representation consumed) and an output (the representation produced), but also something we will refer to as a "meta-output": this provides information about how the translation was performed.

At a high level of abstraction, our architecture will be as follows. Interpretation proceeds as a series of non-deterministic translation steps, each producing a set of possible outputs and associated meta-outputs. The final translation step produces an executable script. The interface attempts to simulate execution of each possible script produced, in order to determine what would happen if that script were selected; simulated execution can itself produce further meta-outputs. Finally, the system uses the meta-output information to decide what to do with the various possible interpretations it has produced. Possible actions include selection and execution of an output script; paraphrasing meta-output information back to the user; or some combination of the two.

In the following section, we present a more detailed description showing how the output/meta-output distinction works in a practical system.

## A Prototype Implementation

The ideas sketched out above have been realized as a prototype spoken language dialogue interface to a simulated version of the Personal Satellite Assistant (PSA). This section gives an overview of the implementation; in the following section, we provide an annotated sample dialogue with the system.

## Levels of Representation

The real PSA [8] is a miniature robot currently being developed at NASA Ames, which is intended for deployment on the Space Shuttle and/or International Space Station. It will be capable of free navigation in an indoor micro-gravity environment, and will provide mobile sensory capacity as a back-up to a network of fixed sensors. The PSA will primarily be controlled by voice commands through a hand-held or head-mounted microphone, with speech and language processing being handled by an offboard processor. Since the speech processing units are not in fact physically connected to the PSA we envisage that they could also be used to control or monitor other environmental functions.

The initial PSA speech interface demo consists of a simple simulation of the Shuttle. State parameters include the PSA's current position, some environmental variables such as local temperature, pressure and carbon dioxide levels, and the status of the shuttle's doors (open/closed). A visual display gives direct feedback on some of these parameters.

The speech and language processing architecture is based on that of the SRI CommandTalk system [6] [13] The system comprised a suite of about 20 agents, connected together using the SRI Open Agent Architecture (OAA) [5]. Speech recognition is performed using a version of the Nuance recognizer [7]. Initial language processing is carried out using the SRI Gemini system [2], using a domain-independent unification grammar and a domain-specific lexicon. The language processing grammar is compiled into a recognition grammar using the methods of [6]; the net result is that only grammatically well-formed utterances can be recognized. Output from the initial language-processing step is represented in a version of Quasi Logical Form [3], and passed in that form to the dialogue manager (DM). We refer to these as *linguistic level* representations.

The aspects of the system which are of primary interest here concern the DM and related modules. Once a linguistic level representation has been produced, the following processing steps occur:

- The linguistic level representation is converted into a *discourse level representation*. This primarily involved regularizing differences in surface form: so for example "measure the pressure" and "what is the pressure?" have different representations at linguistic level, but the same representation at discourse level.

- If necessary, the system attempts to resolve instances of ellipsis and anaphoric reference. For example, if the previous command was "measure temperature at flight deck", then the new command "lower deck" will be resolved to an expression meaning "measure temperature at lower deck". Similarly, if the previous command was "move to the crew hatch", then the command "open it" will be resolved to "open the crew hatch". We call the output of this step a *resolved discourse level representation*.

- The resolved discourse level representation is converted into an executable script. This involves two sub-steps. First, quantified variables are given *scope*: for example, "go to the flight deck and lower deck and measure pressure" becomes something approximately equivalent to the script

    ```
    foreach x (flight_deck lower_deck)
      go_to $x
      measure pressure
    end
    ```

    The point to note here is that the `foreach` has scope over both the `go_to` and the `measure` actions; an alternate (incorrect) scoping would be

    ```
    foreach x (flight_deck lower_deck)
      go_to $x
    end
    measure pressure
    ```

    The second sub-step is to attempt to optimize the plan. In the current example, this can be done by reordering the list `(flight_deck lower_deck)`. For instance, if the PSA is already at the lower deck, reversing the list will mean that the robot only makes one trip, instead of two.

- The final step in the interpretation process is *plan evaluation*: the system tries to work out what will happen if it actually executes the plan. Among other things, this gives the DM the possibility of comparing different interpretations of the original command, and picking the one which is most efficient.

## How Meta-outputs Participate in the Translation

The above sketch shows how context-dependent interpretation is arranged as a series of non-deterministic translation steps; in each case, we have described the input and the output for the step in question. We now go back to the concerns of the previous section, Theoretical Ideas. First, note that each translation step is in general fallible. We give several examples:

- One of the most obvious cases arises when the user simply issues an invalid command, such as requesting the PSA to open a door which is already open. In this case, the plan evaluation step results in the final script issuing an error message as part of its meta-output; the DM can decide to relay this back to the user. Note that plan evaluation does not involve actually executing the final script, which can be important. For instance, if the command is "Go to the crew hatch and open it" and the crew hatch is already open, the interface has the option of informing the user that there is a problem without first carrying out the "go to" action.

- A slightly more complex case involves plan costs. During plan evaluation, the system simulates execution of the output script while keeping track of execution cost. (Currently, the cost is just an estimate of the time required to execute the script). Execution costs are again treated as meta-outputs, and passed back through the interpreter so that the plan optimization step can make use of them.

- The resolution step can give rise to similar kinds of meta-output. For example, a command may include a referring expression that has no denotation, or an ambiguous denotation; for example, the user might say "both decks", presumably being unaware that there are in fact three of them. Once again, the fact that a presupposition failure has occurred is treated formally as a meta-output, and the DM has the possibility of informing the user of their incorrect belief. A particularly interesting case is that of ambiguous denotation; as example 5 in the sample dialogue shows, the DM can react to the meta-output by asking an appropriate clarification question.

## A Compact Architecture for Dialogue Management Based on Scripts and Meta-Outputs

None of the individual functionalities outlined above are particularly novel in themselves. What we find new and interesting is the fact that they can all be expressed in a uniform way in terms of the script output/meta-output architecture. This section presents three examples illustrating how the architecture can be used to simplify the overall organization of the system.

### Integration of Plan Evaluation, Plan Execution and Dialogue Management

Recall that the DM simulates evaluation of the plan before running it, in order to obtain relevant meta-information. At plan execution time, plan actions result in changes to the world; at plan evaluation time, they result in *simulated* changes to the world and/or produce meta-outputs.

Conceptualizing plans as scripts rather than logical formulas permits an elegant treatment of the execution/evaluation dichotomy. There is one script interpreter, which functions both as a script executive and a script evaluator, and one set of rules which defines the procedural semantics of script actions. Rules are parameterized by execution type which is either "execute" or "evaluate". In "evaluate" mode, primitive actions modify a state vector which is threaded through the interpreter; in "execute" mode, they result in commands being sent to (real or simulated) effector agents. Conversely, "meta-information" actions, such as presupposition failures, result in output being sent to the meta-output stream in "evaluate" mode, and in a null action in "execute" mode. The upshot is that a simple semantics can be assigned to rules like the following one, which defines the action of attempting to open a door which may already be open:

```
procedure(
  open_door(D),
  if_then_else(
    status(D,open_closed,open),
    presupp_failure(already_open(D)),
    change_status(D,open_closed,open)))
```

## Using Meta–Outputs to Choose Between Interpretations

As described in the preceding section, the resolution step is in general non-deterministic and gives rise to meta-outputs which describe the type of resolution carried out. For example, consider a command involving a definite description, like "open the door". Depending on the preceding context, resolution will produce a number of possible interpretations; "the door" may be resolved to one or more contextually available doors, or the expression may be left unresolved. In each case, the type of resolution used appears as a meta-output, and is available to the DM when it decides which interpretation is most felicitous. By default, the DM's strategy is to attempt to supply antecedents for referring expressions, preferring the most recently occurring sortally appropriate candidate. In some cases, however, it is desirable to allow the default strategy to be overridden: for instance, it may result in a script which produces a presupposition failure during plan evaluation. Treating resolution choices and plan evaluation problems as similar types of objects makes it easy to implement this kind of idea.

## Using Meta–Outputs to Choose Between Dialogue Management Moves

Perhaps the key advantage of our architecture is that collecting together several types of information as a bag of meta-outputs simplifies the top-level structure of the DM. In our application, the critical choice of dialogue move comes after the DM has selected the most plausible interpretation. It now has to make two choices. First, it must decide whether or not to paraphrase any of the meta-outputs back to the user; for example, if resolution was unable to fill some argument position or find an antecedent for a pronoun, it may be appropriate to paraphrase the corresponding meta-output as a question, e.g. "where do you mean?", or "what do you mean by 'the door'?". Having all the meta-outputs available together means that the DM is able to plan a coherent response: so if there are several meta-outputs which could potentially be worth paraphrasing, it typically realizes only the most important ones.

Second, if interpretation was able to produce a well-formed plan, the DM currently has the three options of executing it, paraphrasing it back to the user as a confirmation question, or doing nothing. Once again, this decision often requires global information about what has happened during the interpretation process. For example, knowing that plan evaluation showed that the plan would take significant time to execute makes it more plausible that the user would prefer to receive a confirmation.

## My Dinner With PSA

This section presents a commented sample dialogue with the implemented PSA interface demo. Text preceded by **USER** represents spoken utterances from the user. Text preceded by **PSA** represents synthesized spoken responses from the PSA. Text inside square brackets describes the actions carried out by the PSA.

1. The PSA can move to various locations and measure environmental factors. In the first exchange, reference resolution is used to interpret the phase "all three decks" as referring to the lower deck, mid deck and flight deck. Plan optimization finds the most efficient way to order the locations, and paraphrases the resulting plan back to the user.
   **USER:** Go to all three decks and measure carbon dioxide.
   **PSA:** I will move to flight deck, mid deck and then lower deck and I will measure carbon dioxide level, okay?
   **USER:** Okay.
   [PSA moves to flight deck]
   **PSA:** The carbon dioxide level at the flight deck is one percent.
   [PSA moves to mid deck]
   **PSA:** The carbon dioxide level at the mid deck is one percent.
   [PSA moves to lower deck]
   **PSA:** The carbon dioxide level at the lower deck is one percent.

2. The next example involves resolution of ellipsis. The relevant context is recovered from the preceding dialogue.
   **USER:** Do the same for the pilot's seat.
   **PSA:** I will move to the pilot's seat and I will measure carbon dioxide level, okay?
   **USER:** Right.
   [PSA moves to pilot's seat]
   **PSA:** The carbon dioxide level at the pilot's seat is one percent.

3. The system encounters a user misconception, which results in a "presupposition failure" meta-output being produced during reference resolution. It consequently decides to alert the user to this, rather than attempt to carry out the command.
   **USER:** Close both doors.
   **PSA:** There are in fact three of them.

4. The user utterance includes a pronoun, "it". The system resolves "it" to "the crew hatch", and issues an appropriate paraphrase in order to confirm that this is what the user meant.
   **USER:** Go to crew hatch and close it.
   **PSA:** I will go to crew hatch and I will close crew hatch, okay?
   **USER:** Yeah.
   [PSA moves to crew hatch and closes it]

5. The user asks the system to close the door, but it is unclear which door they are referring to. An "ambiguous reference" meta-output is produced by reference resolution, resulting in the DM issuing a clarification question.
   **USER:** Close the door.

**PSA:** Which door do you mean?

6. The user response is successfully interpreted by reference resolution as an instance of ellipsis. Trying to close the crew hatch a second time however causes a presupposition failure meta-output to be generated during plan evaluation. The system informs the user of this and takes no action.

   **USER:** The crew hatch.

   **PSA:** The crew hatch is already closed.

7. The user presents three locations in a suboptimal order. Plan evaluation finds a more efficient ordering.

   **USER:** Move to storage lockers, commander's seat and flight deck and measure temperature.

   **PSA:** I will move to flight deck, commander's seat and then storage lockers and I will measure temperature, okay?

   **USER:** Sure.

   [PSA moves to flight deck]

   **PSA:** The temperature at the flight deck is 19.9 degrees Celsius.

   [PSA moves to commander's seat]

   **PSA:** The temperature at the commander's seat is 19.9 degrees Celsius.

   [PSA moves to storage lockers]

   **PSA:** The temperature at the storage lockers is 19.9 degrees Celsius.

8. The user asks the system to repeat the previous command. Since the PSA is now starting from a different position, its most efficient plan is no longer the same, and it informs the user of this.

   **USER:** Do that again.

   **PSA:** I will move to storage lockers, commander's seat and then flight deck and I will measure temperature, okay?

   **USER:** Yes.

   [PSA moves to storage lockers]

   **PSA:** The temperature at the storage lockers is 19.9 degrees Celsius.

   [PSA starts moving to commander's seat]

9. The user interrupts the PSA in the middle of the plan, and asks it to go back to its last stopping place. Since "stop" may involve a potentially hazardous condition, the interface obeys immediately, without attempting to confirm first.

   **USER:** Stop.

   [PSA stops]

   **USER:** Go back.

   [PSA returns to storage lockers]

## Conclusion

We have sketched a generic architecture for spoken language dialogue interfaces to semi-autonomous agents, based on the standard notion of translating to successively deeper levels of representation. The novel aspects of our scheme center around two ideas: firstly, that the final output representations are best conceptualized not as logical expressions but rather as programs in a scripting language; secondly, that steps in the translation process should produce not only a simple output, but also meta-information describing how the output was produced. We have presented examples suggesting how several apparently diverse types of dialogue behavior can be simply captured within our framework, and outlined a prototype implementation of the scheme.

## References

[1] N. Badler, R. Bindiganavale, J. Bourne, J. Allbeck, J. Shi, and M. Palmer. 1999. Real Time Virtual Humans. In *International Conference on Digital Media Futures*.

[2] J. Dowding, J. Gawron, D. Appelt, L. Cherny, R. Moore, and D. Moran. 1993. Gemini: A Natural Language System for Spoken Language Understanding. In *Proceedings of the Thirty-First Annual Meeting of the Association for Computational Linguistics*, Columbus, OH.

[3] J. van Eijck and R. Moore. 1992. Semantic Rules for English. In *The Core Language Engine*, H. Alshawi (ed). MIT Press. Cambridge, MA: London, England.

[4] K. Konolige, K. Myers, E. Ruspini, and A. Saffiotti. 1993. Flakey in Action: The 1992 AAAI Robot Competition. SRI Technical Note 528, SRI, AI Center, SRI International, 333 Ravenswood Ave., Menlo Park, CA 94025.

[5] D. Martin, A. Cheyer, and D. Moran. 1998. Building Distributed Software Systems with the Open Agent Architecture. In *Proceedings of the Third International Conference on the Practical Application of Intelligent Agents and Multi-Agent Technology*, Blackpool, Lancashire, UK. The Practical Application Company Ltd.

[6] R. Moore, J. Dowding, H. Bratt, J. Gawron, Y. Gorfu, and A. Cheyer. 1997. CommandTalk: A Spoken-Language Interface for Battlefield Simulations. In *Proceedings of the Fifth Conference on Applied Natural Language Processing*, pages 1–7, Washington, DC. Association for Computational Linguistics.

[7] Nuance Communications Incorporated. http://www.nuance.com/. As of 10/5/99.

[8] Personal Satellite Assistant Project. http://ic-www.arc.nasa.gov/ic/psa/. As of 10/5/99.

[9] D. Perzanowski, A. Schultz, and W. Adams. 1998. Integrating Natural Language and Gesture in a Robotics Domain. In *IEEE International Symposium on Intelligent Control: ISIC/CIRA/ISAS Joint Conference*, pages 247–252, Gaithersburg, MD: National Institute of Standards and Technology.

[10] D. Perzanowski, A. Schultz, W. Adams, and E. Marsh. 1999. Goal Tracking in a Natural Language Interface: Towards Achieving Adjustable Autonomy. In *ISIS/CIRA99 Conference*, Monterey, CA. IEEE.

[11] D. Pym, L. Pryor, and D. Murphy. 1995. Actions as Processes: A Position on Planning. In *Working Notes, AAAI Symposium on Extending Theories of Action,* pages 169–173.

[12] R. W. Smith. 1997. An Evaluation of Strategies for Selective Utterance Verification for Spoken Natural Language Dialog. In *Proceedings of the Fifth Conference on Applied Natural Language Processing*, pages 41–48.

[13] A. Stent, J. Dowding, J. Gawron, E. Bratt, and R. Moore. 1999. The CommandTalk Spoken Dialogue System. In *Proceedings of the Thirty-Seventh Annual Meeting of the Association for Computational Linguistics*, pages 183–190, College Park, Maryland.

[14] D. Traum and James Allen. 1994. Discourse Obligations in Dialogue Processing. In *Proceedings of the 32nd Annual Meeting of the Association for Computational Linguistics*, pages 1–8.

[15] D. R. Traum and C. F. Andersen, 1999. Representations of Dialogue State for Domain and Task Independent Meta-Dialogue. In *Proceedings of the IJCAI'99 Workshop on Knowledge And Reasoning In Practical Dialogue Systems*, pages 113–120.

[16] B. Webber. 1995. Instructing Animated Agents: Viewing Language in Behavioral Terms. In *Proceedings of the International Conference on Cooperative Multimodal Communication*. Eindhoven Netherlands.

[17] T.A. Winograd. 1973. A Procedural Model of Language Understanding. In *Computer Models of Thought and Language*, R.C. Shank and K.M. Colby, (eds.). Freeman. San Francisco, CA.